\documentclass[letterpaper, 10 pt, journal,twoside]{IEEEtran}  

\usepackage{amsmath}
\usepackage{cite}
\usepackage{amsfonts}
\usepackage{graphicx}
\begin{document}

\title{Optimised Design and Performance Comparison of Soft Robotic Manipulators}

\author{Arnau Garriga-Casanovas$^1$, Shen Treratanakulchai$^1$, Enrico Franco$^1$, Emilia Zari$^1$, Varell Ferrandy$^2$, Vani Virdyawan$^2$, Ferdinando Rodriguez y Baena$^1$

\thanks{$^1$Mechatronics In Medicine Laboratory, Department of Mechanical Engineering, and Hamlyn Centre, Imperial College London, UK. {\tt\small a.garriga-casanovas14@imperial.ac.uk}}
\thanks{$^2$Mechanical Engineering Department, Institut Teknologi Bandung, Kota Bandung, Indonesia. }

}
\maketitle 

\begin{abstract}
Soft robotic manipulators are attractive for a range of applications such as medical interventions or industrial inspections in confined environments. A myriad of soft robotic manipulators have been proposed in the literature, but their designs tend to be relatively similar, and generally offer a relatively low force. This limits the payload they can carry and therefore their usability. A comparison of force of the different designs is not available under a common framework, and designs present different diameters and features that make them hard to compare. In this paper, we present the design of a soft robotic manipulator that is optimised to maximise its force while respecting typical application constraints such as size, workspace, payload capability, and maximum pressure. The design presented here has the advantage that it morphs to an optimal design as it is pressurised to move in different directions, and this leads to higher lateral force. The robot is designed using a set of principles and thus can be adapted to other applications. We also present a non-dimensional analysis for soft robotic manipulators, and we apply it to compare the performance of the design proposed here with other designs in the literature. We show that our design has a higher force than other designs in the same category. Experimental results confirm the higher force of our proposed design.
\end{abstract}

\section{Introduction}
Soft robotic manipulators offer a range of attractive features such as low cost, lack of moving parts which mean that they can be easily miniaturised, flexible transmission lines, compatibility with magnetic-resonance imaging (MRI) and with biological tissue, and inherent safety when interacting with delicate environment, to name a few. This makes them well suited for applications in medicine such as endoluminal surgery, specifically for colorectal surgery or lung interventions \cite{Imperialreview}. They also show promise in the field of industrial inspections to enter confined environments and perform in situ inspections. In this work, we focus on soft robots actuated by a pressurised fluid, typically air, since they represent one of the most common designs.

A myriad of soft robotic manipulators with fluidic actuation have been proposed in the literature, with notable examples \cite{SuzumoriFirst,Stiff-Flop, Stiff-FlopConf,Clemson1,Clemson5,Clemson0,Leuven3, LeuvenPAMsBending,TrivediReview, michigan, SFU}, and minimally invasive surgery has been a common application for soft robots for at least a decade \cite{ReviewSoftMedical,NatureReview,Colobot,RN71}. The majority of existing designs tend to be relatively similar, and orbit around a robot design with a tubular silicone structure and a set of internal chambers that can be pressurised.

The main drawback of these robots is that they have low lateral force, which limits significantly the payloads they can carry and the forces they can apply \cite{Imperialreview}. In this work, we focus on lateral force because it is generally the weakest feature of soft robots.  Axial tension force in these robots is generally related to the stiffness of their silicone and can be increased with stiffer materials in the axial direction, so it is not relevant. Axial compression force can be increased by pressurising all chambers of the manipulator, but that is typically equivalent to a pressurized cylinder made of an inextensible fabric, so it is easy to achieve and not relevant from a design perspective. It should also be noted that, in this work, we primarily consider soft robotic manipulators that offer the capability of bending in any direction in 3D space. This is because soft robots with 1 degree of freedom (DOF) exist and offer relatively high lateral force in the direction of bending, but they offer only passive, low force in the other directions.

\begin{figure}
    \centering
    \includegraphics[width=\columnwidth]{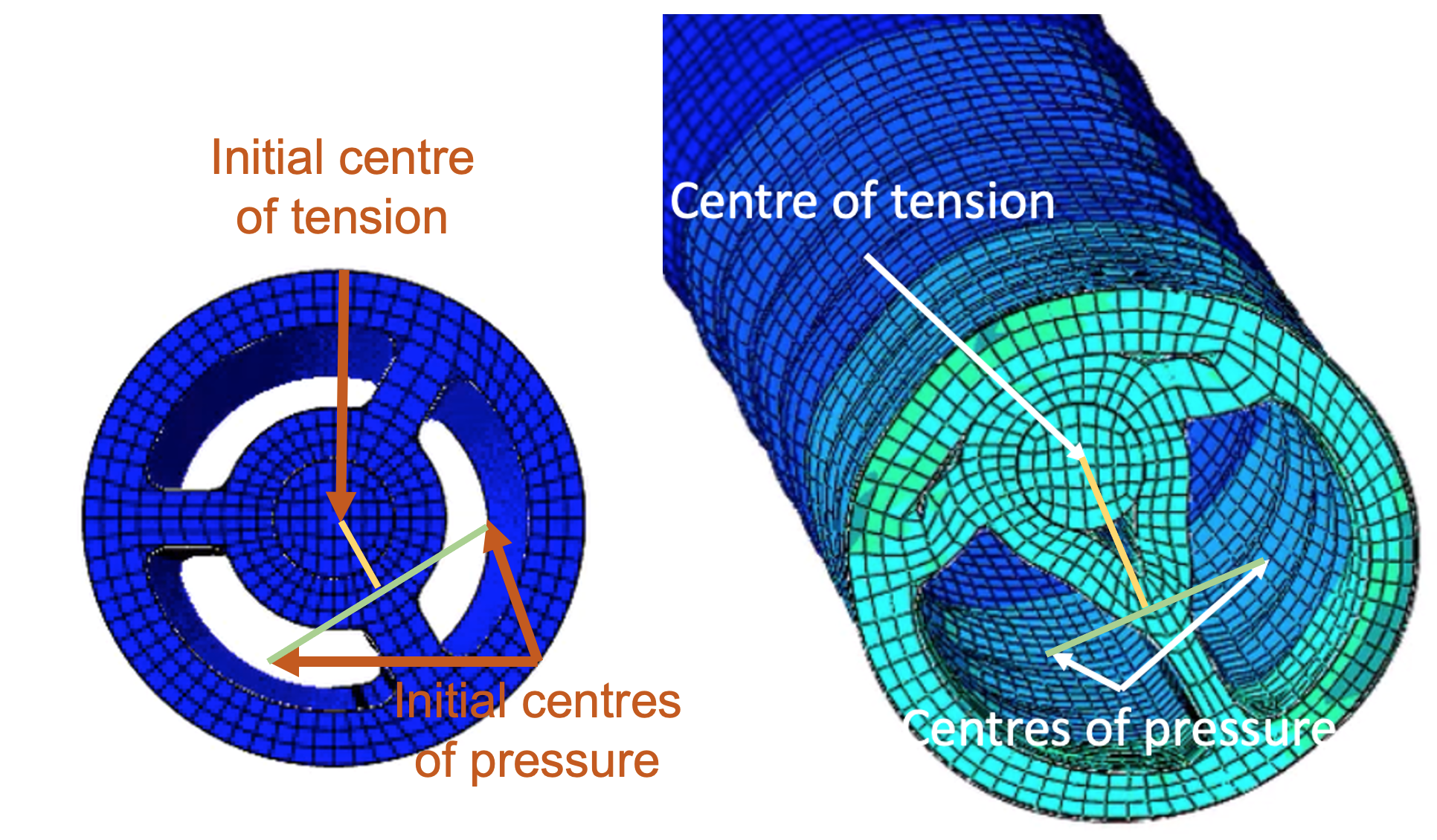}
    \caption{Cross-sections of the proposed morphing design in undeformed and deformed configuration from applying pressure, obtained using finite element simulation in Abaqus. The distance between the equivalent centre of pressures and the centre of tension is shown with a yellow line for both undeformed and deformed cross-section. The increase in that distance when the cross section deforms implies a larger lever arm for the bending moment created by pressure.}
    \label{FigCrossSection}
\end{figure}

A recent comparison of soft robot designs can be found in \cite{Imperialreview}. However, most existing designs have different diameters, which makes it difficult to extrapolate and compare them. In addition, in a number of cases, the specifications and performance of the different designs is not fully reported. It should also be noted that one design in \cite{Imperialreview} stands out as offering a high force, which is Stiff-flop \cite{Stiff-Flop, Stiff-flopReal2}. However, Stiff-flop is 34 mm diameter, it is difficult to miniaturise because of the minimum critical size to enable granular jamming, and it only offers high force when it is rigid, but loses it when it is moving. As a result, it cannot carry a significant payload nor apply forces actively to the environment in the range of maximum force reported.

In this paper, we present the design of a soft robotic manipulator optimised to maximise its force while respecting typical application constraints such as size and miniaturisation potential, mobility, payload carrying capability, and limits on maximum usable pressure (typical in medicine). The design presented here has the advantage that it exploits the internal structural deformation that occurs when it is pressurised, morphing towards an optimal design as it is pressurised to move in different directions, as illustrated in Figure \ref{FigCrossSection}. This leads to higher lateral force, which is employed here as key performance indicator. In order to compare our design with many other ones in the literature, we develop a non-dimensional analysis. We apply this analysis and we show that our design is the strongest in the category of soft robots with pneumatic actuation and similar mobility in terms of DOFs and range. We also present experiments to validate the analysis, and to confirm that our design offers higher force.

The main contributions of the paper are the following: first, the design of a soft robotic manipulator that morphs as it is pressurised, and that is stronger than existing designs; second, the development of a non-dimensional analysis for soft robotic manipulators relying on Buckingham's Pi theorem; third, the experimental comparison of our design and multiple other designs to show which is the strongest and validate the work.

The paper is structured as follows. The design of the soft robotic manipulator proposed in this work is presented in section \ref{Design}, together with a description of its morphing features. Then the non-dimensional analysis for soft robotic manipulators is developed in section \ref{NDA}. The experiments to validate the work are described in section \ref{Experiments}, and their results are reported in section \ref{Results}. Lastly, concluding remarks are described in section \ref{Conclusion}. 

\begin{figure}
    \centering
    \includegraphics[width=\columnwidth]{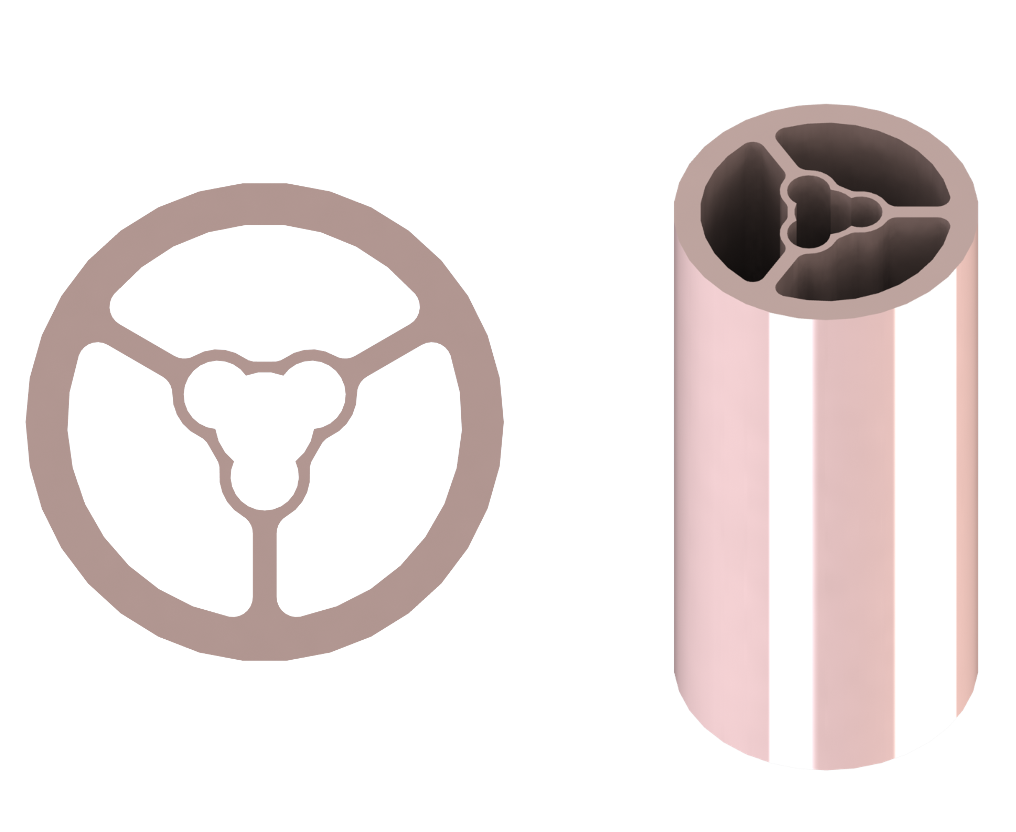}
    \caption{CAD render of the proposed morphing design showing its cross section (left) and a perspective view (right).}
    \label{FigDesignRender}
\end{figure}

\section{Soft robotic manipulator design} \label{Design}

\subsection{Design requirements} \label{Requirements}
The soft robotic manipulator design presented in this work is aimed at a prototypical medical application, since it is one of the most common use cases for soft robots. A soft robotic manipulator for medicine may consist of a set of segments stacked serially to create a robot with a high number of DOFs. Here we focus on the design of one segment of robotic manipulator capable of 2 DOFs, since it is the fundamental building block for a manipulator. However, the design of the robot segment needs to offer the possibility of stacking multiple segments to create a manipulator with more DOFs.

The requirements for the robot segment are defined based on typical requirements for colorectal interventions, and are summarised in Table \ref{TableRequirements}.

The design objective is to maximise the lateral force of the soft robotic manipulator segment so that it can carry a payload as heavy as possible or apply forces on the environment. 

\begin{table}
\centering
\begin{tabular}{rllll} 
\hline
Max. OD & DOFs & Bending & Max. pressure & Payload\\ \hline
12 mm  & 2 per segment  & 90 deg & 3 bar & 2 mm  channel \\ \hline
\end{tabular}
\caption{Design requirements for a segment of soft robotic manipulator.  } \label{TableRequirements}
\end{table}


\subsection{Design principles} \label{Principles}
The design is derived using the design principles in \cite{Garriga-Casanovas2017Design} to maximise its force. The main principles are summarised below for completeness. 

The total cross-section of the device should be maximised to occupy all available room. The area of any internal chambers should be maximised, to ensure that the region of the cross-section corresponding to the pressurised fluid is maximal. The thickness of any internal partition walls separating chambers should be minimised. The longitudinal stiffness in the part of the robot structure that extends when the robot bends should be minimised. The longitudinal stiffness in the part of the robot that does not extend to achieve bending should be maximised. The longitudinal stiffness distribution should be selected so that the maximum stiffness is concentrated near the edge of the cross section, near side toward which the robot bends. The geometry of the cross-section should be designed so that the distance between the centre of pressures and the centre of structural tension forces in the cross section is maximised.

\subsection{Proposed design}
\begin{figure}
    \centering
    \includegraphics[width=\columnwidth]{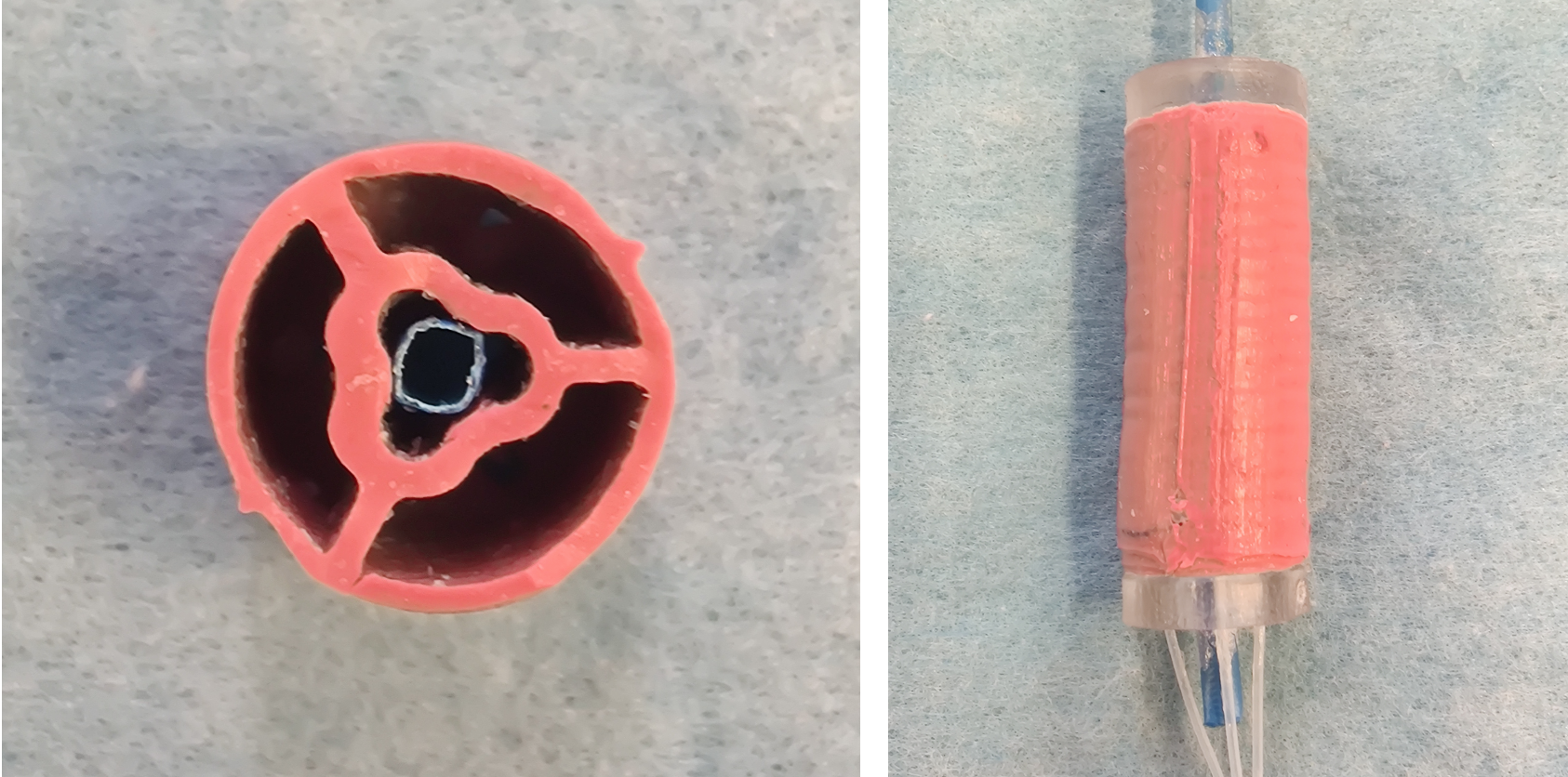}
    \caption{Manufactured segment of the proposed showing its cross section (left), and a side view of the segment (right).}
    \label{FigDesignMade}
\end{figure}

The proposed design was derived by applying the design principles in subsection \ref{Principles} to the design requirements in \ref{Requirements}. The resulting design is shown in Figure \ref{FigDesignRender}. It consists of a tubular structure made of silicone (Elastosil 4601M) with three chambers in the cross section in order to be able to bend in any direction. Using more than three chambers would lead to a larger part of the cross section being occupied by partition walls, which reduces the area available for the chambers, and is thus undesirable as it would leads to lower force. The centre of the design has an inextensible tube that absorbs the majority of the axial tension when one or more chambers are pressurised, and thereby acts as the counterbalance for the pressure force, creating a bending moment. The inextensible central tube also offers a working channel, which in this case is 2 mm ID. The partition walls have a minimal thickness in order to maximise chamber area, and to facilitate cross-sectional deformation. In this work, the partition walls are 0.8 mm thick which is the limit that could be achieved in manufacturing without leakage. The outer wall of the design has an inextensible fibre wrapped around it in a double helix arrangement to prevent radial expansion while enabling longitudinal extension and thereby bending, as in standard soft robots. The ideal design would consist of a pleated outer wall that minimises bending resistance, but manufacturing such an outer wall at the scale required here remains an unsolved challenge, so a smooth outer wall with helical fibres is used instead. A manufactured segment of the design together with its cross section are shown in Figure \ref{FigDesignMade}.

The design proposed may resemble existing soft robots cosmetically since it has a tubular silicone structure with three chambers. However, it differs significantly since it is designed to morph as it is pressurised, and its features, such as chamber geometry, deformable partition walls, inextensible central tube that can bend and absorbs structural tension, are engineered to tend to an optimal design. When a differential pressure is applied in the chambers, the device is engineered to facilitate cross-sectional deformation. The result is a significant cross-sectional deformation when pressure is applied, as shown in Figure \ref{FigCrossSection} (right), which was obtained by simulating the pressurisation of the robot design shown in \ref{FigCrossSection} (left) in two chambers using finite element simulation software Abaqus, and following the simulation setup described in \cite{Garriga-Casanovas2017Design}. In particular, the pressurised chambers occupy practically all of the cross-section, and the inextensible central tube is fully displaced to the side opposing the applied pressure, which leads to a desirable configuration that offers significantly higher bending force. This is because the bending moment generated by the pressure is roughly proportional to the product of pressure times chamber area times the distance between the centre of pressures and the centre of tension forces in the structure. In this case, the main stiff element in the structure is the inextensible central tube, which absorbs the majority of axial tension, and thus it acts as the centre of tension forces in the structure. As a result, the distance between the centre of pressures and centre of tension forces in the structure is nearly equal to the radius, which is nearly equal to the maximum available. Similarly, the area of the pressurised chambers occupies practically the entire cross-section, which is also close to the maximum available. Thus, the configuration is nearly optimal and maximises force. 

This near optimal configuration remains near optimal when bending in any direction. Since the cross-section deforms when a differential pressure is applied, and the inextensible central tube is always pushed to the side opposing the applied pressure, the robot always morphs to an optimal configuration regardless of the direction in which it needs to bend.

It should also be noted that, at the two ends of the robot segment, the inextensible central tube is attached to the centre of the segment. This means that, when a differential pressure is applied and the tube is pushed to one side of the robot, it would need to follow a longer path between the two ends of the segment to remain attached to them, which creates additional tension in the inextensible tube. This behavior is similar to that found in a bow when it is tensioned, with the difference that in the soft robot the pressure applies the force that displaces the inextensible tube, whereas in a bow it is a person that pulls the string from the bow frame. The result however is similar, as the tension in the inextensible central tube of the soft robotic manipulator also contributes further to the bending force. 



The improvement of the design proposed here with respect to designs that have constant cross-section, such as Stiff-flop, is significant from a theoretical perspective. The chamber area in this design can reach nearly double the size when pressurised with respect to the original size, as shown in Figure \ref{FigCrossSection}. In addition, the distance between the centre of pressures and the centre of structural tension also nearly doubles when the cross-section is pressurised, as also highlighted using a yellow line in Figure \ref{FigCrossSection}. Both of these elements contribute to the bending moment, which is the product of both of them times pressure, so the result can be expected as a bending force that could theoretically reach four times the force of a similar design with fixed cross-section. The experimental evaluation of force is presented in section \ref{Results}.

The improvement of the design proposed here with respect to traditional designs such as the flexible microactuator (FMA) \cite{SuzumoriFirst,SuzumoriSecond} is also significant. In general, soft robots can be considered to achieve bending either via an asymmetry in geometry or in stiffness. In the case of the FMA, it is mainly designed to achieve bending via an asymmetry in geometry (even if some cross-sectional deformation may occur accidentally). In addition, the FMA can "waste" some of the applied differential pressure in extending its structure instead of bending because it has a relatively low axial stiffness throughout its structure. However, the design proposed here employs both the asymmetry in geometry and in stiffness thanks to the morphing in the cross-section when it is pressurised, leading to higher performance. Moreover, the design proposed here has the inextensible central tube which absorbs most of the axial tension, and ensures that practically all of the applied pressure translates into bending.

\begin{figure}
    \centering
    \includegraphics[width=\columnwidth]{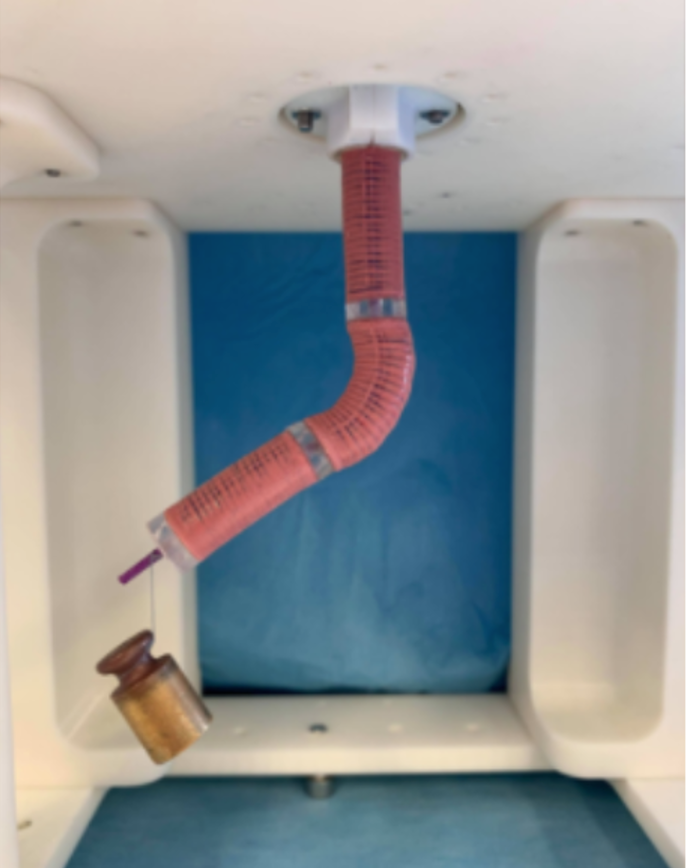}
    \caption{Full prototype of the soft robotic manipulator with 12 mm OD made of three segments, which have 2 DOF each, resulting in a 6 DOF robotic manipulator.}
    \label{FigPrototype}
\end{figure}

The design presented here consists of one segment with 2 DOF, and a manufactured prototype is shown in Figure \ref{FigDesignMade}. Multiple segments can be stacked, and a 6 DOF robot can be created with three segments, as in the prototype shown in Figure \ref{FigPrototype}. Miniature tubes of 0.3 mm OD can be used to pressurise the distal segments, and these can be passed through the central working channel or the small channels next to it shown in Figure \ref{FigDesignMade} (left), so practically no changes are needed to the design of the individual segments to create the full robot shown in \ref{FigPrototype}.

\section{Non-dimensional analysis} \label{NDA}

\subsection{Theoretical formulation}
A non-dimensional analysis is developed in this section for soft robotic manipulators actuated by a pressurised fluid, relying on Buckingham's Pi theorem. According to Buckingham's Pi theorem, the deformation of the manipulators considered in this work depends on a set of non-dimensional groups, as long as they are geometrically equivalent, under equivalent external load distribution, and pressurisation in one chamber or multiple chambers with a given ratio of pressures between chambers. These non-dimensional groups can be selected to be $\frac{p}{E_r}$, $\frac{F}{E_rd^2}$, $\frac{E_f}{E_r}$, $\frac{E_c}{E_r}$, $\frac{E_p}{E_r}$, as well as a parameters $\nu_r$, $\nu_f$, $\nu_c$, and $\nu_p$, where $p$ is the reference magnitude of the pressure applied, $F$ is the magnitude of the external wrenches applied for a given wrench distribution that generally excludes gravitational forces since they are typically negligible, $d$ is a reference length corresponding to the diameter of the device, $E_r$ is a coefficient related to the stiffness of the rubber (which can correspond to the Young's modulus or the $c_{10}$ coefficient in a Neo-Hookean constitutive law), $E_f$ is the stiffness of the outer fibres, $E_c$ is the stiffness of the central tube, $E_p$ is the stiffness of the partition walls in designs where these are made of a different material than the outer wall, and $\nu_r$, $\nu_f$, $\nu_c$, and $\nu_p$ are the Poisson ratios of the rubber, outer fibres, central rod and partition walls, respectively. The condition of geometric equivalence for soft robotic manipulators in the non-dimensional analysis implies that the aspect ratio between diameter and length is maintained, and thus only one dimensional variable $d$ is necessary to specify size. It should be noted that here the use of the magnitude $p$ in the non-dimensional groups imposes the stress at the surfaces where pressure is applied, and is equivalent to the use of a reference magnitude of stress in the structure of the device. 

In the deformation of the soft robotic manipulator, it can be assumed that the outer fibres are inextensible. In addition, the effect of $\nu_f$, $\nu_c$, and $\nu_p$ on the behaviour of the complete device can be assumed to be negligible in general. Moreover, rubbers used as soft material are generally incompressible, so $\nu_r$ is generally constant. Thus, the non-dimensional groups of $\frac{E_f}{E_r}$, $\nu_f$, $\nu_c$, $\nu_r$ and $\nu_p$ are considered to have a negligible effect or be irrelevant, and the dominant non-dimensional groups are reduced to $\frac{p}{E_r}$, $\frac{F}{E_rd^2}$, $\frac{E_c}{E_r}$, and $\frac{E_p}{E_r}$. Also, in some designs such as the one described in section \ref{Design}, the central rod can be assumed to be practically inextensible, and in other devices such as the FMA the central rod is not present. In these cases, the group $\frac{E_c}{E_r}$ is not relevant. Finally, in designs where the partition walls must also be made of the same material as the outer wall due to fabrication constraints, such as the design in section \ref{Design}, the non-dimensional groups are reduced to $\frac{p}{E_r}$, $\frac{F}{E_rd^2}$.

Using this non-dimensional analysis, the behaviour of geometrically equivalent devices with either different size, different material, different magnitude of external wrenches, or different applied pressure can be expected to be equivalent if the non-dimensional groups are equal. In addition, the number of variables influencing the behaviour of a soft robotic manipulator in the non-dimensional case is reduced by two relative to the dimensional case, which simplifies the study of these devices. This analysis can also be generalised to designs with additional regions or fibres with different stiffness by adding non-dimensional groups corresponding to the ratio of stiffnesses in the different regions.

\subsection{Application}
This non-dimensional analysis enables the extrapolation of the behaviour of a soft robotic manipulator to different sizes, pressures, loads, and materials. This leads to a set of relevant insights.

\begin{table*}[t]
\centering
\begin{tabular}{rllll} 
\hline
Material & Ultimate stress [MPa] & Ultimate strain & Shore hardness & $c_{10}$ coeff. [kPa] \\ \hline
DragonSkin 10  & 2.75 & 663$\%$ & 10A & 42.5 \\ 
DragonSkin 20  & 3.8  & 620$\%$ & 20A & n/a \\ 
DragonSkin 30  & 3.45 & 384$\%$ & 30A & n/a \\ 
Elastosil M4601 & 6.5  & 700$\%$ & 28A & 131.2 \\
Ecoflex OO-30  & 1.38 & 900$\%$ & OO30 & 12.7 \\ 
Ecoflex OO-50  & 2.17 & 980$\%$ & OO50 & 25 \\ \hline
\end{tabular}
\caption{Material properties of available hyperelastic rubbers relevant to create deformable structures in soft robotic manipulators.  } \label{TableMaterials}
\end{table*}

A first insight of interest is related to the non-dimensional group $\frac{F}{E_rd^2}$. This elucidates that the external wrenches that can be supported by a device with specified material properties and a given deformation varies with the square of the diameter of the device. This implies that scaling up a soft robotic manipulator increases the capability of supporting forces as the square of the diameter.

Another relevant insight is that, for devices with equal material properties and no external loads, the relation between applied pressure and deformation is constant regardless of the device scale. This implies that the pressure required for operation is determined by the design and desired deflection, but not by the scale of the robot, which is particularly relevant when considering medical applications.

In terms of materials, the non-dimensional analysis indicates that an equal deformation is achieved in designs made of materials with different stiffness if the pressure and external wrenches are varied proportionally to maintain the non-dimensional groups constant. This suggests that stiffer materials lead to robots capable of supporting higher wrenches with an equal deformation, which is generally desirable, provided that pressure is also increased. 

The stress in a stiffer material also increases proportionally in the case of equivalent deformation. Thus, the use of stiff materials with a relatively low ultimate stress can be undesirable. However, for materials with similar ultimate strain, which is the case of some of the hyperelastic rubbers used as soft materials in robotic manipulators with fluidic actuation, the non-dimensional analysis indicates that the use of stiffer materials is desirable as it leads to devices that can support higher wrenches, provided that the assumptions in the previous subsection are satisfied. This guides the selection of an appropriate silicone for the robot, as described in the next subsection.

\subsection{Material selection}
The hyperelastic material of which the structure of the soft robot is made affects the performance, and needs to be selected carefully. The main alternatives for hyperelastic rubbers in the literature and offered by suppliers are summarised in Table \ref{TableMaterials}. The material properties were obtained from the literature. There are two potential values in the literature for the elastic coefficient $c_{10}$ for Elastosil 4601M, from \cite{Elastosil2} and \cite{Elastosil1}. In \cite{Elastosil2} it is implied that $c_{10}=131.2 \pm 24.3 kPa$. Instead, in \cite{Elastosil1} they characterise Elastosil 4601M using a Yeoh law with $c_{10}=110 kPa$ and $c_{20}=20 kPa$, which can be used to approximate a Neo-Hookean law with $c_{10}=110 kPa$ for relatively low strains. In this work, we use the average value from \cite{Elastosil2} since it is the average from a set of tests, but it should be noted that some variation of up to $18\%$ is possible in practice.

The materials in Table \ref{TableMaterials} present significant differences in their stiffness and ultimate strain. The selection of material can be made with aid from the non-dimensional analysis presented in the previous subsection. As can be seen in Table \ref{TableMaterials}, the ultimate strain of Elastosil M4601, by Wacker Chemie AG (Munich, Germany), is higher than that of DragonSkin 10, 20, and 30, by Smooth-On Inc. (Macungie, US). In addition, the stiffness of Elastosil M4601 is also higher or comparable to that of these other materials. Hence, according to the non-dimensional analysis in the previous section \ref{NDA}, Elastosil M4601 is preferable over DragonSkin 10, 20 or 30, since it can reach equivalent deformations to DragonSkin 10 or 20 while supporting higher wrenches, and at the deformation where DragonSkin 30 reaches its ultimate strain, Elastosil M4601 can still increase its strain by further increasing pressure, leading to higher support of wrenches. 

When comparing the ultimate strain of Elastosil M4601 with that of Ecoflex OO-30 and OO-50, by Smooth-On Inc., it can be seen that the ultimate strain of the former is somewhat lower. This means that a device made of Ecoflex OO-30 or OO-50 can reach an equal deformation as a device made of Elastosil M4601 at the point where the latter reaches its ultimate strain, and then keep increasing deformation and thus resistance to wrenches by increasing pressure. However, the stiffness of Ecoflex OO-30 or OO-50 is at least three times lower that of Elastosil M4601. According to the non-dimensional analysis in section \ref{NDA}, this implies that at the maximum strain of Elastosil M4601, with an equivalent deformation, the wrenches that a device made of Elastosil M4601 can support are at least three times those that a design made of Ecoflex OO-30 or OO-50 can support. At the point of a strain of 700$\%$, devices made of Ecoflex OO-30 or OO-50 can be further pressurised to increase strain up to 900$\%$, and thus increase somewhat the wrenches they can support. However, the additional increase in wrenches that can be support when increasing pressure and strain up to 900$\%$ is not expected to be triple the wrenches that can be supported at 700$\%$ strain. Thus, the wrenches that a device made of Elastosil M4601 can support at its maximum strain of 700$\%$ are expected to be higher than those that a device made of Ecoflex OO-30 or OO-50 can support at its maximum strain of 900$\%$. As a result, Elastosil M4601 is the most suitable material to support the highest wrenches.

\section{Experiments} \label{Experiments}
Experiments were conducted to validate the non-dimensional analysis and to measure the force of the design proposed in this paper. 

\begin{figure}
    \centering
    \includegraphics[width=\columnwidth]{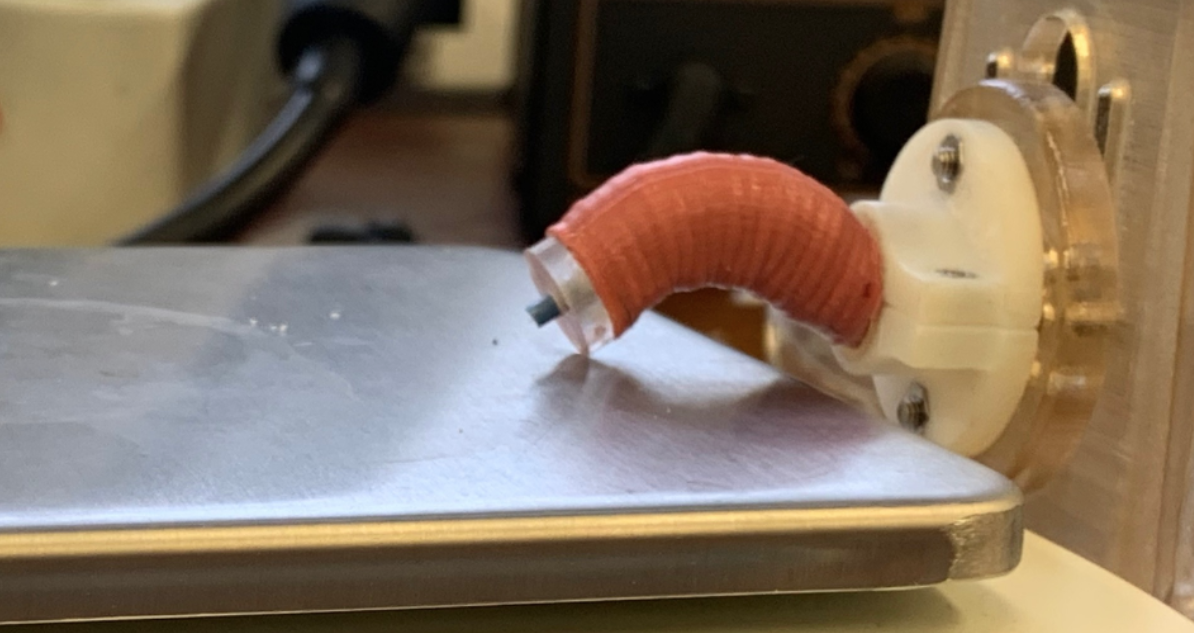}
    \caption{Experimental setup used to measure the lateral force of the soft robotic manipulator.}
    \label{FigForceExperiment}
\end{figure}

\subsection{Experiment description}
The experimental set-up is shown in Figure \ref{FigForceExperiment}. It consists of a holder to keep the base of the robot horizontal, a robot segment attached to it, and a scale under the segment to measure the lateral force that it applies. The chambers of the robot are pressurised using supply tubes, which are connected to a set of pressure regulators (PRE1-U08, AirCom Pneumatic GmbH,Ratingen, Germany), and controlled from a laptop via a microcontroller (mbed nxp lpc1768) via a serial link. This set-up was used to measure the force of two robot segments made of two different silicone rubbers with different stiffnesses. 


\subsection{Results and non-dimensional analysis validation}
\begin{figure}
    \centering
    \includegraphics[width=\columnwidth]{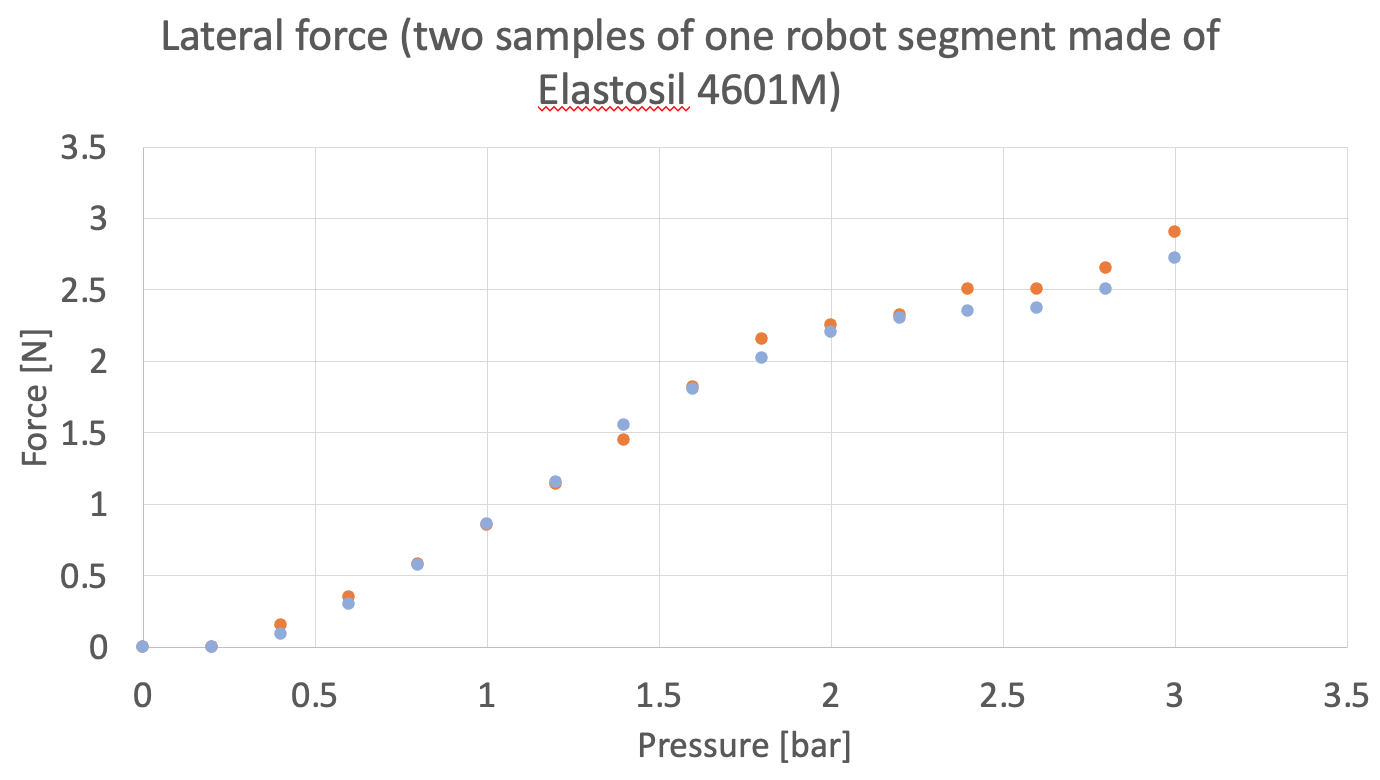}
    \caption{Lateral force from a robot segment with the optimal design, 12 mm OD, and made of Elastosil 4601M.}
    \label{FigForceElastosil}
\end{figure}

The results of force measured as a function of pressure for a segment made of Elastosil 4601M with 12 mm OD are shown in Figure \ref{FigForceElastosil}. As can be seen, the segment can reach forces of 2.9N at 3 bar, at which pressure the segment begins to fail. The tests were repeated using two equal segments made of the same material to confirm the repeatability of the experiment, both reported in Figure \ref{FigForceElastosil}. The variation in force between the two equal segments was less than 10$\%$.

\begin{figure}
    \centering
    \includegraphics[width=\columnwidth]{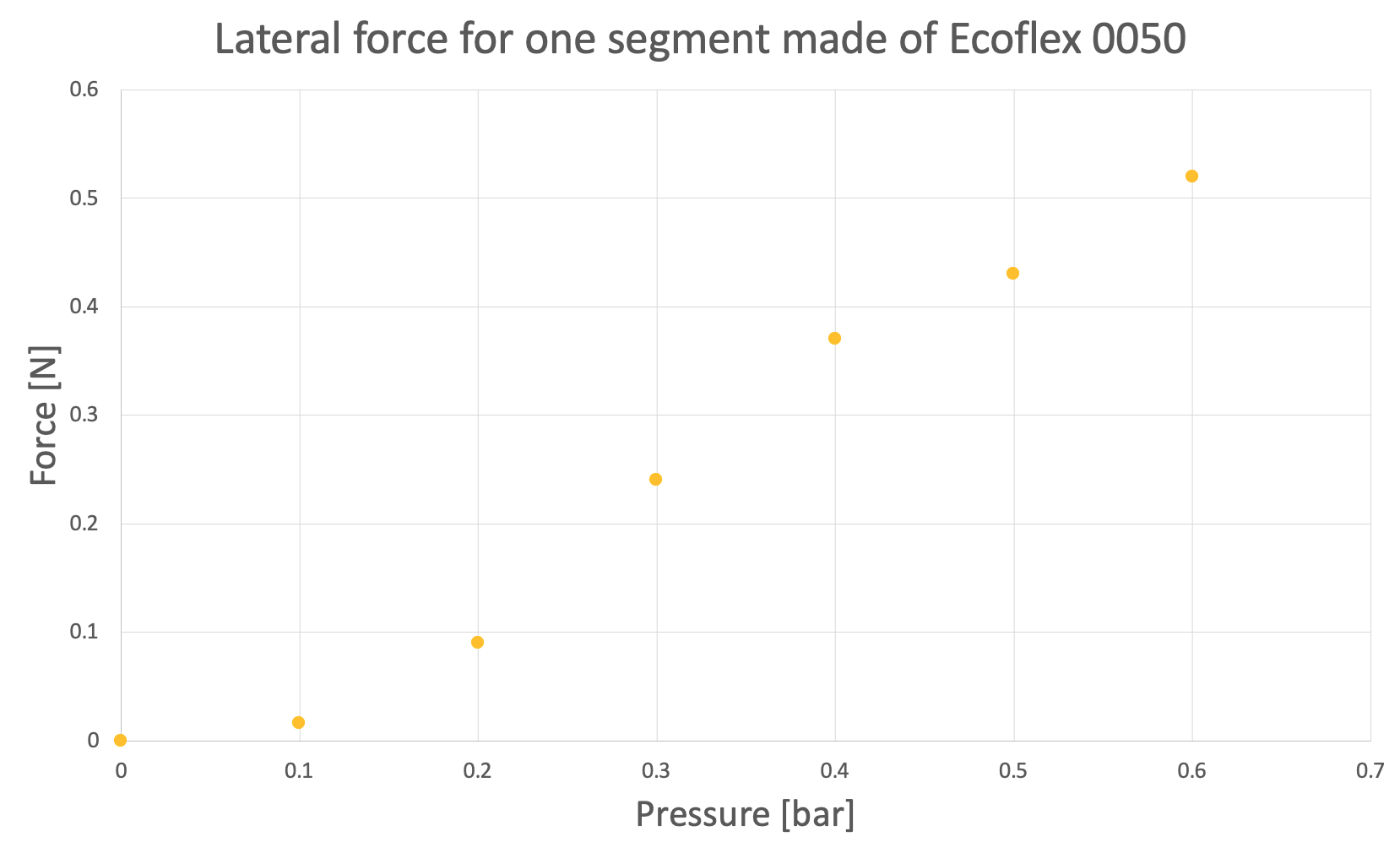}
    \caption{Lateral force from a robot segment with the optimal design, 12 mm OD, and made of Ecoflex 0050.}
    \label{FigForceEcoflex}
\end{figure}

The results of force of the same segment design and same dimension (12 mm OD) but made of Ecoflex 0050 are shown in Figure \ref{FigForceEcoflex}. The experiment was conducted up to a pressure where the robot began to experience failure. The segment can reach a force of 0.52N at 0.6 bar, which was observed to be the failure pressure for this material.

The ratio between the Neo-hookean stiffness coefficient of the materials used, Elastosil 4601M and Ecoflex 0050, is 5.2 according to Table \ref{TableMaterials}. Comparing the results in Figures \ref{FigForceElastosil} and \ref{FigForceEcoflex}, there is a scaling factor in terms of force and pressure between the results corresponding to the segments made of the two different materials of approximately 5. This is similar to the ratio of Neo-hookean stiffness coefficient of the two materials. 

\begin{figure}
    \centering
    \includegraphics[width=\columnwidth]{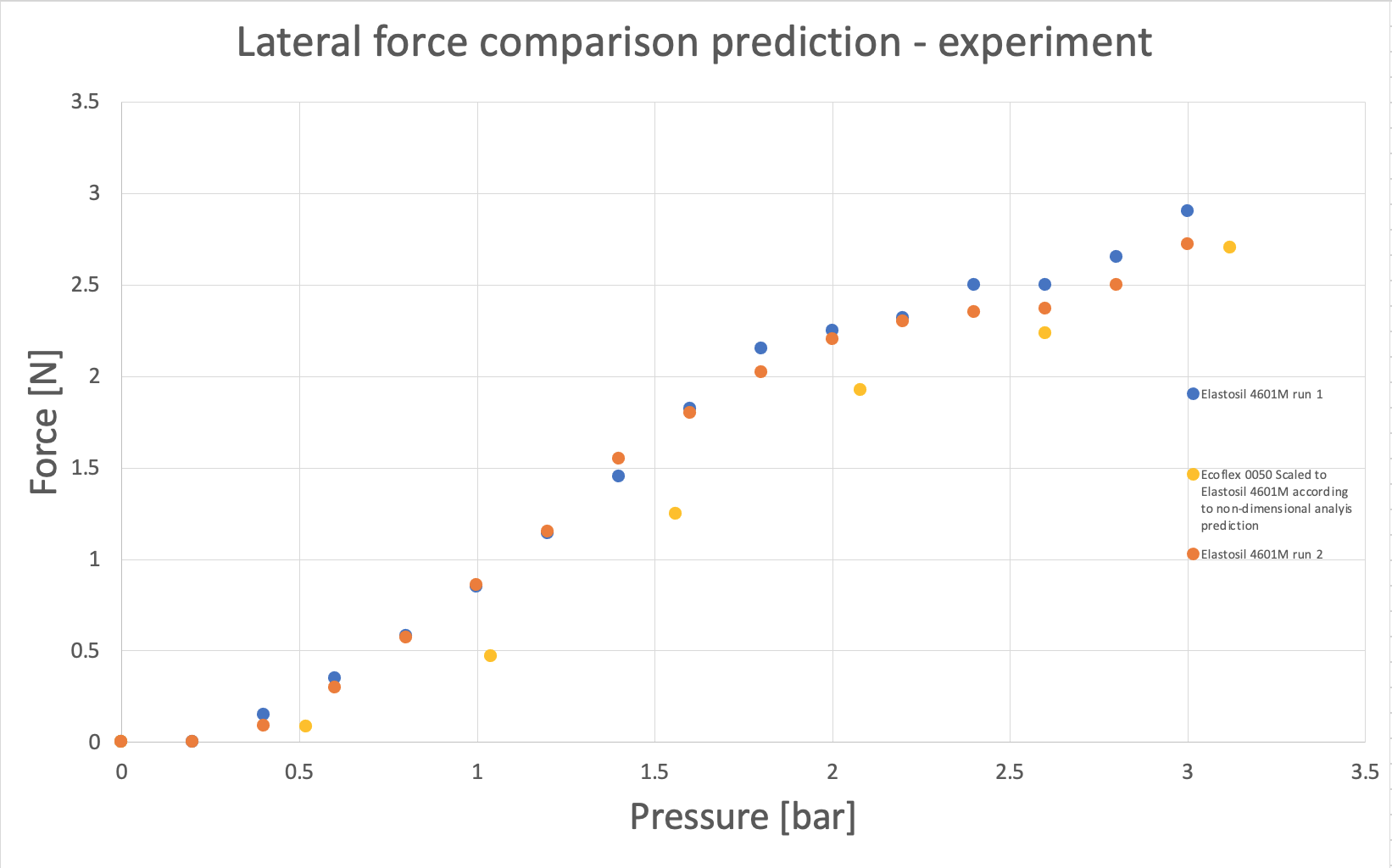}
    \caption{Comparison of lateral force predicted by non-dimensional analysis scaling of Ecoflex 0050 to Elastosil 4601M, and experimental measurement of lateral force of device made of Elastosil 4601M.}
    \label{FigForceComparison}
\end{figure}

The theoretical lateral force of a device made of Ecoflex 0050 scaled using the non-dimensional analysis to a theoretical device made of Elastosil 4601M using the ratio of Neo-hookean stiffness coefficient between the two devices of 5.2 is plotted in Figure \ref{FigForceComparison}. The same Figure also shows the experimental measurement of force for the devices made of Elastosil 4601M for comparison. The both plots in Figure \ref{FigForceComparison} show a relatively good agreement. This confirms that the results from one segment made of one material can be extrapolated to an equal segment made of a different material by using the scaling corresponding to the ratio of stiffnesses and maintaining the non-dimensional groups descried in section \ref{NDA}, and matches the behavior predicted by the non-dimensional analysis.

It should be noted that the scaling is not perfect, and the trends in the graphs in Figure \ref{FigForceComparison} differ to some extent. This can be attributed both to experimental inaccuracies in manufacturing the segments and to the fact that the inextensible central tube used in all the segments made of different materials was the same and was not scaled according to the non-dimensional groups in section \ref{NDA}. This means that the tube bending stiffness is more dominant for the softer material, Ecoflex 0050, particularly at lower pressures where the force of the robot is lower. This effect is less important at higher pressures where the bending stiffness of the central tube is less dominant compared to the bending force generated by pressure. Thus, the comparison of the scaling between both graphs in Figure \ref{FigForceComparison} is more relevant in the region corresponding to higher pressures used in each test.



\subsection{Force measurement of proposed design}
The experiments show the force of the proposed design as a function of pressure for different materials. Given that the maximum pressure for the medical application considered in this work is 3 bar, the material Elastosil 4601M is confirmed as a suitable choice, since other materials exhibit failure at lower force and pressures. 

The maximum force of one segment of the design proposed here is approximately 2.9N, as can be seen in Figure \ref{FigForceElastosil}. This implies that the force of a full prototype made of three segments such as that shown in Figure \ref{FigPrototype}, which is three times the length and thus any lateral force at the tip applies a triple moment at the base, is a third of that of a single segment, or approximately 0.97N.

\section{Performance comparison} \label{Results}
The performance of the design proposed in this work is compared in this section with that of other relevant existing designs reported in the literature in the same category of soft robotic manipulators actuated by a pressurised fluid that could meet the design requirements in section \ref{Requirements}. It should be noted that the literature in various cases does not include values of lateral force, as reported in \cite{Imperialreview}. Here we include the designs where lateral force is reported, following \cite{Imperialreview} with the addition of recent published work. 

As previously mentioned, the designs in the literature all have different diameters. The non-dimensional analysis is first used to extrapolate their force to a chosen outer diameter of 12 mm. Table \ref{TableComparison} shows the maximum force of the main existing designs in the literature when scaled to a 12 mm OD.

\begin{table}
\centering
\begin{tabular}{rl} 
\hline
Robot design & Max. lateral force at 12 mm OD [N]  \\ \hline
Tsinghua University \cite{TsinghuaForce} & 0.384  \\
University of Hong Kong \cite{HongKong} & 0.82  \\
Stiff-flop design in \cite{Stiff-flopSim} & 0.09 \\
Stiff-flop jamming in \cite{Stiff-Flop, Stiff-flopReal2} & 0.84 \\
Our design & 2.9 \\ \hline
\end{tabular}
\caption{Maximum lateral force for one 2DOF segment of main designs in the literature when scaled to a outer diameter of 12mm, and maximum force of the design proposed in this work at the same diameter.} 
\label{TableComparison}
\end{table}

It should be noted that the force values for the stiff-flop design reported in \cite{Stiff-Flop, Stiff-flopReal2} correspond to lateral force when the robot is locked in a rigid configuration using granular jamming, and an external force is applied on it. However, the robot cannot actively move, and thus this force cannot be used to carry payloads or actively apply forces on the environment. In \cite{Stiff-Flop, Stiff-flopReal2}, it is also reported that the stiff-flop module can reach forces of 47.1N in a 33 mm OD robot (equivalent to 6.2N in a 12 mm OD). However, the authors in \cite{Stiff-Flop, Stiff-flopReal2} report that these forces are measured by placing a load cell on top of a vertical robot and pressurising it, and thus are understood to correspond to extension forces rather than lateral forces. The values for extension forces are not used here since this paper focuses on lateral force. Extension forces are typically significantly higher than lateral forces. For example, in \cite{TsinghuaForce}, extension forces are nearly an order of magnitude higher than lateral forces. 

As can be seen in Table \ref{TableComparison}, the lateral force of all designs reported in the literature is below 1N when scaled to a 12 mm OD. The design proposed in this work offers a higher force of 2.9N. This is 200$\%$ above the existing highest lateral force. 

One factor that affects lateral force is the length of the robot, as already mentioned previously. As such, shorter designs with a smaller aspect ratio will typically have comparatively higher lateral force. This could in some cases bias the comparison of force to some extent. However, shortening a design typically means reducing the bending angle, which is typically proportional to the length of the design. Thus, to achieve the desired performance of 90 degree bending selected in this work, which is common across the literature, robots cannot be designed to be as short as possible, and instead a design compromise needs to be struck. As a result, the comparison of designs based on lateral force as in Table \ref{TableComparison} for designs that can typically reach bending of 90 degrees remains valid  for most cases.

\section{Conclusions} \label{Conclusion}
A soft robotic manipulator design optimised to maximise its force was presented in this work. A non-dimensional analysis was also developed to compare the performance of the proposed design with other designs in the literature. The non-dimensional analysis was validated with experiments, and insights were extracted from it. In particular, the non-dimensional analysis was used to extrapolate the force of different designs in the literature to a given diameter. The force of existing designs was compared with the design proposed in this work, and it was shown that the design proposed here offers the highest force. 

The design proposed in this work offers a force, size, mobility, and payload capabilities that are suitable for medical applications. The design exploits internal structural deformation to maximise force by morphing its internal features into an optimal configuration when pressurised to bend in each desired direction. In addition, it is derived using a set of design principles that make it possible to easily adapt it to the requirements of each application. In future work, we plan to apply this design on a fully functioning 6DOF robotic manipulator for MIS.

\section{Acknowledgements}
This work was supported by the British Council under the Newton Fund Institutional Links, grant number 623531377.

\bibliography{bibDesign}

\begin{thebibliography}{10}
\providecommand{\url}[1]{#1}
\csname url@samestyle\endcsname
\providecommand{\newblock}{\relax}
\providecommand{\bibinfo}[2]{#2}
\providecommand{\BIBentrySTDinterwordspacing}{\spaceskip=0pt\relax}
\providecommand{\BIBentryALTinterwordstretchfactor}{4}
\providecommand{\BIBentryALTinterwordspacing}{\spaceskip=\fontdimen2\font plus
\BIBentryALTinterwordstretchfactor\fontdimen3\font minus
  \fontdimen4\font\relax}
\providecommand{\BIBforeignlanguage}[2]{{%
\expandafter\ifx\csname l@#1\endcsname\relax
\typeout{** WARNING: IEEEtran.bst: No hyphenation pattern has been}%
\typeout{** loaded for the language `#1'. Using the pattern for}%
\typeout{** the default language instead.}%
\else
\language=\csname l@#1\endcsname
\fi
#2}}
\providecommand{\BIBdecl}{\relax}
\BIBdecl

\bibitem{Imperialreview}
M.~Runciman, A.~Darzi, and G.~P. Mylonas, ``{Soft Robotics in Minimally
  Invasive Surgery 1},'' vol.~6, no.~4, 2019.

\bibitem{SuzumoriFirst}
K.~Suzumori, S.~Iikura, and H.~Tanaka, ``{Development of flexible microactuator
  and its applications to robotic mechanisms}.''\hskip 1em plus 0.5em minus
  0.4em\relax IEEE International Conference on Robotics and Automation (ICRA),
  1991.

\bibitem{Stiff-Flop}
M.~Cianchetti, T.~Nanayakkara, T.~Ranzani, G.~Gerboni, K.~Althoefer,
  P.~Dasgupta, and A.~Menciassi, ``{Soft Robotics Technologies to Address
  Shortcomings in Today's Minimally Invasive Surgery: The STIFF-FLOP
  Approach},'' \emph{Soft Robotics}, vol.~1, no.~2, pp. 122--131, 2014.

\bibitem{Stiff-FlopConf}
J.~Czarnowski, M.~Macia, J.~G{\l}{\'{o}}wka, M.~Cianchetti, and A.~Menciassi,
  ``{New STIFF-FLOP module construction idea for improved actuation and
  sensing},'' in \emph{IEEE International Conference on Robotics and Automation
  (ICRA)}, 2015, pp. 2901--2906.

\bibitem{Clemson1}
W.~McMahan, V.~Chitrakaran, M.~Csencsits, D.~Dawson, I.~D. Walker, B.~Jones,
  M.~Pritts, D.~Dienno, M.~Grissom, and C.~D. Rahn, ``{Field trials and testing
  of the OcotArm continuum manipulator},'' in \emph{IEEE international
  conference on robotics and automation (ICRA)}, 2006.

\bibitem{Clemson5}
I.~S. Godage, R.~Wirz, I.~D. Walker, and R.~J. Webster, ``{Accurate and
  Efficient Dynamics for Variable-Length Continuum Arms: A Center of Gravity
  Approach},'' \emph{Soft Robotics}, vol.~2, no.~3, pp. 96--106, 2015.

\bibitem{Clemson0}
I.~D. Walker, D.~M. Dawson, T.~Flash, F.~W. Grasso, R.~T. Hanlon, B.~Hochner,
  W.~M. Kier, C.~C. Pagano, C.~D. Rahn, and Q.~M. Zhang, ``{Continuum robot
  arms inspired by cephalopods},'' \emph{SPIE Conference on Unmanned Ground
  Vehicle Technology}, vol. 5804, pp. 303--314, 2005.

\bibitem{Leuven3}
G.~Smoljkic, G.~Borghesan, D.~Reynaerts, J.~D. Schutter, J.~V. Sloten, and
  E.~V. Poorten, ``{Hybrid robotic system for applications in robotic
  surgery},'' \emph{Proceedings of the 5th Joint Workshop on New Technologies
  for Computer/Robot Assisted Surgery}, pp. 1--3, 2015.

\bibitem{LeuvenPAMsBending}
A.~Devreker, B.~Rosa, A.~Desjardins, E.~J. Alles, L.~C. Garcia-Peraza,
  E.~Maneas, D.~Stoyanov, A.~L. David, T.~Vercauteren, J.~Deprest, S.~Ourselin,
  D.~Reynaerts, and E.~{Vander Poorten}, ``{Fluidic actuation for
  intra-operative in situ imaging},'' in \emph{IEEE International Conference on
  Intelligent Robots and Systems}, vol. 2015-Decem, 2015, pp. 1415--1421.

\bibitem{TrivediReview}
D.~Trivedi, C.~D. Rahn, W.~M. Kierb, and I.~D.Walkerc, ``{Soft robotics:
  Biological inspiration, state of the art, and future research},''
  \emph{Applied Bionics and Biomechanics}, 2008.

\bibitem{michigan}
J.~Bishop-Moser, G.~Krishnan, C.~Kim, and S.~Kota, ``{Design of soft robotic
  actuators using fluid-filled fiber-reinforced elastomeric enclosures in
  parallel combinations}.''\hskip 1em plus 0.5em minus 0.4em\relax IEEE
  International Conference on Intelligent Robots and Systems (IROS), 2012.

\bibitem{SFU}
B.~C.-M. Chang, J.~Berring, M.~Venkataram, C.~Menon, and M.~Parameswaran,
  ``{Bending fluidic actuator for smart structures},'' \emph{Smart Materials
  and Structures}, vol.~20, no.~3, p. 035012, 2011.

\bibitem{ReviewSoftMedical}
A.~{De Greef}, P.~Lambert, and A.~Delchambre, ``{Towards flexible medical
  instruments: Review of flexible fluidic actuators},'' \emph{Precision
  Engineering}, vol.~33, no.~4, pp. 311--321, 2009.

\bibitem{NatureReview}
D.~Rus and M.~T. Tolley, ``{Design, fabrication and control of soft robots},''
  \emph{Nature}, vol. 521, no. 7553, pp. 467--475, 2015.

\bibitem{Colobot}
G.~Chen, M.~T. Pham, T.~Maalej, H.~Fourati, R.~Moreau, and S.~Sesmat, ``{A
  Biomimetic steering robot for Minimally invasive surgery application},''
  \emph{IN-TECH}, 2009.

\bibitem{RN71}
B.~Zhang, C.~Hu, P.~Yang, Z.~Liao, and H.~Liao, ``Design and modularization of
  multi-dof soft robotic actuators,'' \emph{IEEE Robotics and Automation
  Letters}, vol.~4, no.~3, pp. 2645--2652, 2019.

\bibitem{Stiff-flopReal2}
M.~Cianchetti, T.~Ranzani, G.~Gerboni, I.~D. Falco, C.~Laschi, S.~Member, and
  A.~Menciassi, ``{STIFF-FLOP Surgical Manipulator : mechanical design and
  experimental characterization of the single module},'' pp. 3576--3581, 2013.

\bibitem{Garriga-Casanovas2017Design}
A.~Garriga-Casanovas, I.~Collison, and F.~R.~y. Baena, ``{Towards a Common
  Framework for the Design of Soft Robotic Manipulators with Fluidic
  Actuation},'' \emph{Soft Robotics}, 2017.

\bibitem{SuzumoriSecond}
K.~Suzumori, S.~Iikura, and H.~Tanaka, ``{Applying a Flexible Microactuator to
  Robotic Mechanisms}.''\hskip 1em plus 0.5em minus 0.4em\relax IEEE
  International Conference on Robotics and Automation (ICRA), 1992.

\bibitem{Elastosil2}
G.~Alici, T.~Canty, R.~Mutlu, and V.~Sencadas, ``{Modeling and Experimental
  Evaluation of Bending Behavior of Soft Pneumatic Actuators Made},'' vol.~5,
  no.~1, pp. 24--35, 2018.

\bibitem{Elastosil1}
B.~Mosadegh, P.~Polygerinos, C.~Keplinger, S.~Wennstedt, R.~F. Shepherd,
  U.~Gupta, J.~Shim, K.~Bertoldi, C.~J. Walsh, and G.~M. Whitesides,
  ``{Pneumatic Networks for Soft Robotics that Actuate Rapidly}.''

\bibitem{TsinghuaForce}
B.~Zhang, C.~Hu, P.~Yang, Z.~Liao, and H.~Liao, ``{Design and Modularization of
  Multi-DoF Soft Robotic Actuators},'' \emph{IEEE Robotics and Automation
  Letters}, vol.~4, no.~3, pp. 2645--2652, 2019.

\bibitem{HongKong}
K.-h. Lee, D.~K.~C. Fu, M.~C.~W. Leong, M.~Chow, and H.-c. Fu, ``{Nonparametric
  Online Learning Control for Soft Continuum Robot : An Enabling Technique for
  Effective Endoscopic Navigation},'' vol.~4, no.~4, pp. 324--337, 2017.

\bibitem{Stiff-flopSim}
Y.~Elsayed, A.~Vincensi, C.~Lekakou, T.~Geng, and C.~M. Saaj, ``{Finite Element
  Analysis and Design Optimization of a Pneumatically Actuating Silicone
  Module},'' vol.~1, no.~4, pp. 255--262, 2014.

\end{thebibliography}

\end{document}